\documentclass{sigkddExp}

\usepackage{booktabs}
\usepackage{graphicx}
\usepackage{tabularx}
\usepackage{array}
\usepackage{amsmath}
\usepackage{amssymb}
\usepackage{xcolor}
\usepackage{enumitem}
\usepackage{tikz}
\usetikzlibrary{positioning,arrows.meta,fit,calc,backgrounds}
\usepackage[T1]{fontenc}
\usepackage{times}
\usepackage{url}
\usepackage{microtype}
\definecolor{navy}{HTML}{1F3A5F}
\definecolor{steel}{HTML}{4A6C8C}
\definecolor{mist}{HTML}{EEF2F6}
\definecolor{sand}{HTML}{F5F1E8}
\definecolor{slate}{HTML}{5B6573}
\definecolor{rust}{HTML}{9A4A3A}

\newcolumntype{Y}{>{\raggedright\arraybackslash}X}
\newcolumntype{P}[1]{>{\raggedright\arraybackslash}p{#1}}
\setlist{nosep,leftmargin=*}
\makeatletter
\usepackage{etoolbox}
\patchcmd{\@sect}{\vskip -14pt}{\vskip -6pt}{}{\typeout{sect patch failed}}
\makeatother
\newcommand{\para}[1]{\par\smallskip\noindent\textbf{#1}\ }
\newcommand{\pos}[1]{%
  \par\medskip\noindent
  \begin{tikzpicture}
    \node[
      fill=mist,
      draw=steel!40,
      line width=0.5pt,
      rounded corners=2.5pt,
      inner xsep=8pt,
      inner ysep=6pt,
      text width=\dimexpr\columnwidth-17pt\relax,
      align=justify,
      font=\small
    ] (box) {#1};
    \draw[navy, line width=2.6pt, line cap=round]
      ([xshift=1.3pt,yshift=-2pt]box.north west) -- ([xshift=1.3pt,yshift=2pt]box.south west);
  \end{tikzpicture}%
  \par\medskip
}
\newcommand{\gapbox}[1]{%
  \par\medskip\noindent
  \begin{tikzpicture}
    \node[
      fill=sand,
      draw=rust!35,
      line width=0.5pt,
      rounded corners=2.5pt,
      inner xsep=8pt,
      inner ysep=6pt,
      text width=\dimexpr\columnwidth-17pt\relax,
      align=justify,
      font=\small
    ] (box) {#1};
    \draw[rust, line width=2.6pt, line cap=round]
      ([xshift=1.3pt,yshift=-2pt]box.north west) -- ([xshift=1.3pt,yshift=2pt]box.south west);
  \end{tikzpicture}%
  \par\medskip
}

\newcommand{\authoremailfont}{\normalfont\sffamily\fontsize{9}{11}\selectfont}

\begin{document}

\title{Toward Knowledge-Centric AI: Discovering, Internalizing, and Using Explicit Knowledge Across Agents and Generations}

\author{
\begin{minipage}{\dimexpr\textwidth-2\tabcolsep\relax}
\centering
{\normalfont\sffamily\fontsize{12}{16}\selectfont
Yisen Gao\textsuperscript{1}, Haoyu Huang\textsuperscript{1},
Zhongwei Xie\textsuperscript{1}, Yufei Li\textsuperscript{1}\\
Hong Ting Tsang\textsuperscript{1}, Boyi Zhang\textsuperscript{1},
Qiao Xiao\textsuperscript{1}\\
Tianshi Zheng\textsuperscript{1}, Jiaxin Bai\textsuperscript{2},
Yangqiu Song\textsuperscript{1}\endgraf}
\vspace{6pt}
{\normalfont\sffamily\fontsize{10}{12}\selectfont
\textsuperscript{1}Department of Computer Science and Engineering\\
The Hong Kong University of Science and Technology, Hong Kong\\
\textsuperscript{2}Department of Computer Science, Hong Kong Baptist University, Hong Kong\endgraf}
\vspace{5pt}
{\authoremailfont
\{ygaodi, hhuangcp, zxiebk, ylivm,\\
httsangaj, bzhangdd, qxiaoac, tzhengad\}@connect.ust.hk\\
baijiaxin@hkbu.edu.hk,\quad yqsong@cse.ust.hk\endgraf}
\end{minipage}
}

\maketitle

\begin{abstract}
Knowledge is how understanding accumulates across people and generations: durable records that outlive their producers and that others can check, combine, teach, and correct. AI agents are now built and run by many independent users and organizations, and each agent is upgraded, replaced, or retired within months. Yet what an agent learns stays trapped in model parameters, a transient context window, or a system-specific memory store that is appended to and retrieved from rather than processed. We argue that the next generation of AI should be knowledge-centric, built around an explicit, continuously evolving knowledge layer independent of any single agent. In this loop, agents discover knowledge from their environments and trajectories; the layer internalizes it by turning fragmented, heterogeneous records and experience into composable, revisable understanding; and agents across systems and generations use the result and return new outcomes for revision. While discovery and use have advanced rapidly with large language model agents, internalization remains the bottleneck. We argue that internalization should become a persistent object of training and maintenance—centered on condition-preserving conceptualization of events and deductive, inductive, and abductive reasoning across noisy, heterogeneous representations (text, tables, graphs, rules, and programs). To power this layer, we call for knowledge foundation models that provide shared neural capacity for knowledge computation, and outline a research agenda showing how an explicit knowledge layer delivers reuse, cost efficiency, auditability, and continuity across agent generations.

\end{abstract}

\section{Introduction}
\label{sec:intro}

Humans do not pass their brains to one another or to the next generation. What is passed on is knowledge in explicit form: laws, theorems, recipes, maps, procedures, encyclopedias, and the scientific literature. Knowledge accumulates along two dimensions. Across contemporaries, people in different places, with different goals and different tools, contribute findings that others test, combine, and build on, usually without working on the same task. Across generations, a physics student today does not rediscover mechanics from observation but reads a compressed and verified account of it, and then extends or corrects it. Explicit records are the medium of both. They are selective, since what is written down is what a community judged worth keeping, and they are durable, inspectable, and composable. Organizational theory describes the same process within firms, where tacit know-how is externalized, combined with other explicit knowledge, and internalized again by new members \cite{nonaka1994dynamic}.

AI agents now face both dimensions. An agent consists of a model and a harness, meaning the prompts, tools, memory, and control loop around the model, and it interacts with an environment \cite{yao2023react,shinn2023reflexion,wang2023voyager,gao2026safe}. In the same period, many independent agent systems run side by side. Different users, teams, and organizations operate their own agents or multi-agent systems, with different models, tools, goals, permissions, and ways of representing what they know. They share no context and no controller, yet one system's finding could inform another's decision. Over time, agents are upgraded, replaced, and retired. A new generation of agents should not have to repeat all the exploration of the previous one, and it should not inherit the previous generation's limitations along with its findings.

Much of what agents learn arrives as experience, in trajectories of actions, environment events, feedback, and outcomes, much as organizations accumulate operation records and consumer services accumulate the event histories of their users. What an agent learns currently ends up in one of three places below, and none of them supports accumulation along both dimensions. In model parameters it becomes implicit: it cannot be read, cited, or moved to a different model, and studies of language models show that parametric knowledge can be directional \cite{berglund2024reversal}, hard to extract unless it was seen in diverse forms during pretraining \cite{allenzhu2024physics31}, and hard to update without side effects \cite{cohen2024ripple}. In a context window it is expensive, used unevenly across long inputs \cite{liu2024lost}, and lost when the session ends. In an external memory store it persists \cite{packer2023memgpt,chhikara2025mem0}, but the store usually belongs to one system and is shaped by its model and harness, and it is mostly appended to and retrieved from rather than abstracted, repaired, and reasoned over as a body of knowledge.

\pos{\textcolor{navy}{\textbf{Position.}} The next generation of AI should be knowledge-centric. It needs an explicit knowledge layer that is independent of any single agent and that keeps evolving. Agents of many systems (1) discover knowledge from their environments and trajectories, (2) the layer internalizes it, turning fragmented and heterogeneous knowledge and experience into understanding that can be composed, reasoned with, and revised, and (3) agents of other systems and later generations use the result. Discovery and use are advancing quickly. Internalization is the bottleneck. It should become a persistent object of training and maintenance, and knowledge foundation models should provide shared neural capacity for it, with knowledge and its computation, not a data format or a task, as their foundation.}

\begin{figure*}[t]
\centering
\includegraphics[width=0.98\textwidth]{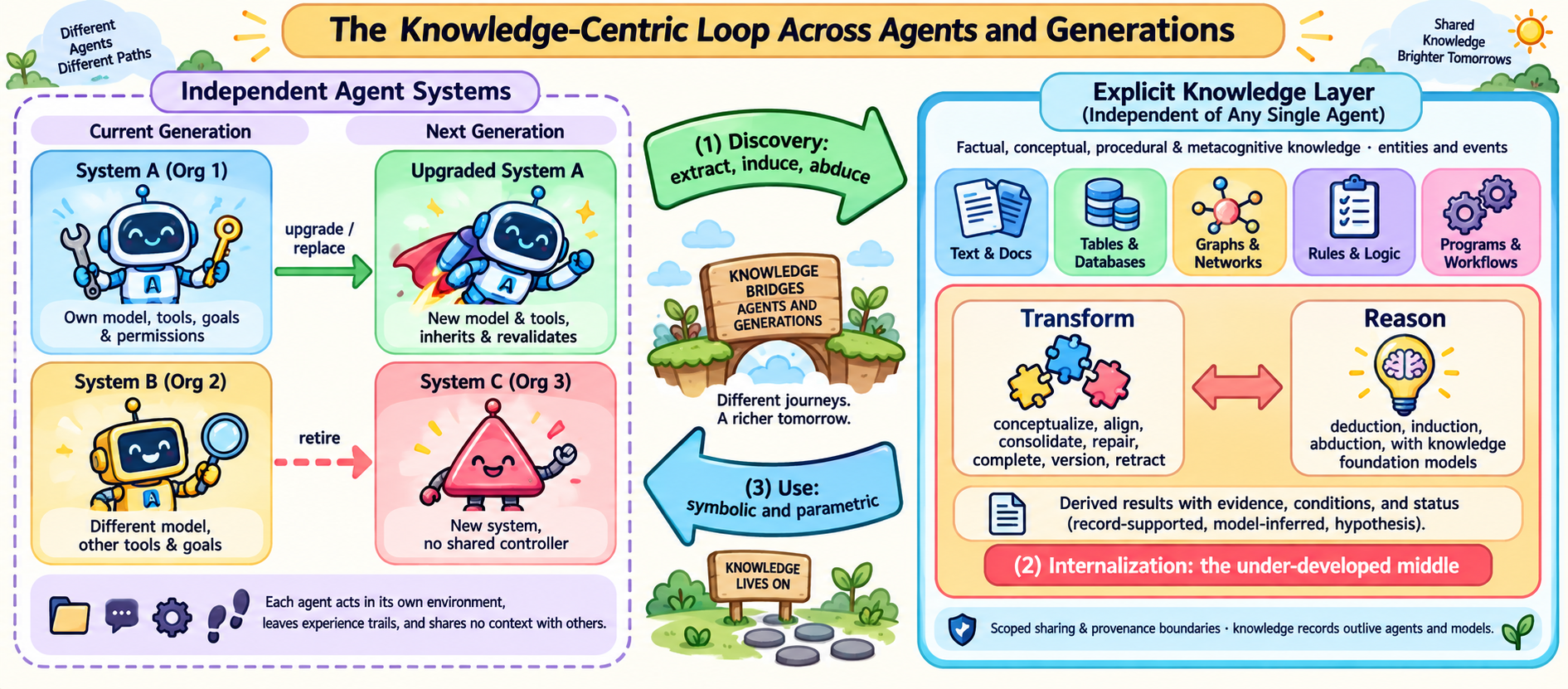}
\caption{The knowledge-centric loop across agents and generations. Independent agent systems, built on different models and run by different organizations, discover knowledge and contribute it to an explicit knowledge layer. The layer is internalized by transformations and reasoning, and the results are used by other systems and by later generations, which inherit, revalidate, and revise them.}
\label{fig:loop}
\end{figure*}

\medskip
The position rests on three principles, which also say when a system counts as knowledge-centric.

\begin{enumerate}
\item[\textbf{P1}] \textbf{Separability.} Knowledge is stored apart from the agents and models that produced it, in units that record content, conditions of validity, provenance, and status (\S\ref{sec:knowledge}). Explicit records should not be interpretable only by the model that wrote them, and model parameters should not be the only authoritative copy of knowledge.
\item[\textbf{P2}] \textbf{Active, evidence-grounded internalization.} Rather than passively storing records, the knowledge layer actively aligns, abstracts, composes, and reasons over knowledge to build and reorganize a coherent body of understanding, while preserving roles, applicability conditions, exceptions, and links to evidence as it repairs and versions the result (\S\ref{sec:internal}).
\item[\textbf{P3}] \textbf{Accumulation across agents and generations.} The goal of the layer is that knowledge produced by one agent helps other agents, built on other models and operated by other parties, and helps later generations, which inherit it by understanding, using, revalidating, and revising it rather than by copying old records.
\end{enumerate}

\para{Relation to prior positions.} Calls to unify language models and knowledge graphs \cite{pan2024unifying,zhu2023kgllm} concern how two families of models can strengthen each other. Chen's perspective on large knowledge models argues for models engineered to manage diverse knowledge structures and names the disentangling of knowledge bases from language models as a key challenge \cite{chen2024lkm}. A recent large knowledge model for science organizes the literature into persistent, source-grounded reasoning graphs that grow incrementally and that many agents retrieve from \cite{huang2026lkm}. We share their commitment to separability (P1). We add three things. First, we treat the stage between storage and use as a distinct object of study, with its own operators, invariants, and objective, and we place condition-preserving conceptualization and three forms of reasoning at its center (P2). Second, we frame the knowledge layer by accumulation across independent systems and across generations of agents and models (P3), which changes what the layer must record and what its models must not monopolize. Third, we identify asymmetries between language and non-language foundation models in both discovery and use, and argue that a knowledge foundation model can build on both.

\para{Contributions.} The contributions of this paper can be summarized as:
\begin{enumerate}
\item \textbf{A formulation of knowledge and the knowledge-centric loop across agents and generations.} We separate explicit knowledge from raw observations, trajectory experience, and model priors by requiring each unit to carry conditions, provenance, and status, and formulate the loop of discovery, internalization, and use with a working objective and a bottleneck hypothesis.
\item \textbf{A systematization of knowledge internalization and knowledge foundation models.} We establish internalization as a persistent object of training and maintenance, centered on cross-carrier alignment, condition-preserving conceptualization, operator invariants, and deductive, inductive, and abductive reasoning, and define knowledge foundation models (KFMs) as shared capacity for this computation.
\item \textbf{A gap analysis across foundation model modalities and a falsifiable research agenda.} We identify key asymmetries between language and structured-data foundation models in both extracting and using explicit knowledge, and lay out concrete research challenges alongside a critical baseline comparison against agents operating directly on raw records.
\end{enumerate}

\section{What Is Knowledge?}
\label{sec:knowledge}

A position about knowledge-centric AI needs a usable notion of knowledge. We do not attempt a philosophical definition. We aim for distinctions that change how systems should be built.

\subsection{From data to knowledge}

The data--information--knowledge--wisdom hierarchy, despite its many variants, captures a widely shared intuition that knowledge is information that has been organized and made actionable \cite{rowley2007wisdom}. For AI systems we adopt a working definition with three parts. Knowledge is information that (i) is organized so that it can be related to other knowledge, (ii) is reusable beyond the situation in which it was observed, and (iii) has been judged worth keeping for use by someone other than the immediate producer. A single fact such as ``paper $P$ appeared at venue $V$'' satisfies these conditions without any abstraction. Abstraction is what distinguishes conceptual knowledge, which is induced from many instances, from such reusable facts.

Three kinds of material take part in knowledge computation without being knowledge by default. Observations, such as the rows of a measurement table or a sensor log, record what was seen. Experience, such as an agent's action trajectory, records what was done and what followed. Model priors and judgments, such as a language model's belief or a forecasting model's prediction, record what a model expects. All three can support knowledge, and knowledge can be derived from them, but the content of a carrier is not automatically knowledge. A log of every action an agent took is experience. The observation that a particular sequence of actions reliably books a flight on a particular website, stated so that another agent can use it, is knowledge. A model's confident prediction is a judgment, and treating it as an observed fact is an error that a knowledge layer should prevent.

Agent trajectories deserve a closer look, because they are a large and fast-growing source of experience. Many of their components are eventualities (\S\ref{sec:types}): an action the agent took, an event in the environment, feedback from a user or a tool, and the outcome that followed. A trajectory is not merely a set of such events. It also records their order, the context in which each occurred, and dependencies among them, such as which observation prompted which action, and some of its content, such as raw tool outputs or page snapshots, need not be represented as eventualities at all. The distinction among records, experience, and knowledge applies directly. The raw log is a record. The trajectory read as what was tried and what followed is experience. A statement such as ``when a checkout request times out after payment is submitted, check the order history before retrying,'' together with the conditions under which it holds and the trajectories that support it, is knowledge. The same reading applies to operation records in organizations and to the event histories of individual users (\S\ref{sec:settings}).

We represent a unit of explicit knowledge as
\begin{equation}
\kappa = (c,\; f,\; C,\; \pi,\; \rho),
\label{eq:unit}
\end{equation}
where $c$ is the content, $f$ is the form in which it is stored (\S\ref{sec:forms}), $C$ is the set of conditions under which it is asserted to hold (a population, period, environment, or task family, together with known exceptions), $\pi$ is its provenance (sources, producing agent and model, supporting observations, and derivation from other units), and $\rho$ is its status (confidence, validation evidence, whether it is observed, inferred, or hypothesized, and version). Conditions and provenance are often omitted in practice. We include them because knowledge without conditions cannot be safely reused elsewhere, and knowledge without provenance cannot be audited, corrected, or retired. Conditions also record dependence on a particular model or tool when there is one, so that a later generation can tell knowledge about the environment from advice about how an old agent should operate. The tuple is a record format, not a guarantee. Whether $C$, $\pi$, and $\rho$ remain correct after knowledge is merged, abstracted, or revised depends on the operators that act on it, which is why \S\ref{sec:invariants} states invariants for those operators.

\subsection{Types and granularity of knowledge}
\label{sec:types}

The revised taxonomy of educational objectives distinguishes four types of knowledge: factual, conceptual, procedural, and metacognitive \cite{krathwohl2002revision}. Cognitive science draws a related line between declarative knowledge, which can be stated, and procedural knowledge, which is exercised in skilled performance and is acquired by compiling declarative knowledge into procedures \cite{anderson1982acquisition}. We adopt the four-type taxonomy because each type maps onto a recognizable class of AI artifacts (Table~\ref{tab:types}).

\begin{itemize}
\item \textbf{Factual knowledge} consists of discrete facts about entities and events, such as ``Paris is the capital of France''. In AI it appears as knowledge graph triples, database rows, and memory entries.
\item \textbf{Conceptual knowledge} consists of categories, schemas, principles, and models that relate facts to each other. Taxonomies of concepts \cite{wu2012probase}, ontologies \cite{gruber1993translation}, governing equations \cite{brunton2016sindy}, causal graphs \cite{zheng2018notears}, logical rules \cite{galarraga2013amie}, and formulaic signals in finance \cite{tang2025alphaagent} are conceptual knowledge.
\item \textbf{Procedural knowledge} is knowledge of how to do something, including methods, algorithms, and criteria for choosing among them. Skill libraries of executable code \cite{wang2023voyager}, reusable workflows \cite{wang2024awm}, and synthesized APIs \cite{zheng2025skillweaver} are procedural knowledge.
\item \textbf{Metacognitive knowledge} is knowledge about cognition and about knowledge itself, such as which strategy tends to work in which situation and how reliable a piece of knowledge is. Insights distilled from past successes and failures \cite{zhao2024expel,ouyang2026reasoningbank}, evolving strategy playbooks \cite{zhang2026ace}, and the conditions $C$ and status $\rho$ of Equation~\eqref{eq:unit} are metacognitive.
\end{itemize}

Orthogonal to type is granularity. Most large knowledge graphs describe entities and their attributes, whereas much of what agents need concerns eventualities, that is, activities, states, and events and the relations among them, such as what typically happens after an event or why it happens. ASER showed that eventuality knowledge can be extracted at scale, with 194 million unique eventualities and 64 million edges among them mined from more than 11 billion tokens of text \cite{zhang2020aser}. Eventualities are not limited to happenings in the external world. The actions, environment events, feedback, and outcomes in agent trajectories, the operations recorded by enterprise systems, and the events of a user's life are eventualities too, so one representation covers both knowledge about the world and experience gathered while acting in it. Granularity matters for internalization because abstraction behaves differently at the two levels. An entity is abstracted by mapping it to a concept, while an eventuality, or a fragment of related eventualities, is abstracted by mapping its arguments to concepts while keeping the roles they play and the relations among its parts (\S\ref{sec:concept}).

Two further observations follow. First, much of what recent agent research calls ``memory'' is metacognitive or procedural rather than factual, which matters because these types need different internalization operations. Second, the types differ widely in maturity. Factual knowledge is extracted and stored at scale, conceptual knowledge is extracted from structured and observational data mainly in commonsense, scientific, and financial niches, and metacognitive knowledge is usually kept as free text without conditions or validation.

\begin{table*}[t]
\centering
\caption{Four types of knowledge \protect\cite{krathwohl2002revision} and their counterparts in AI systems.}
\label{tab:types}
\vspace{4pt}
\footnotesize
\renewcommand{\arraystretch}{1.15}
\begin{tabularx}{\textwidth}{@{}P{2.0cm} Y Y Y P{2.8cm}@{}}
\toprule
\textbf{Type} & \textbf{Description} & \textbf{Examples in AI systems} & \textbf{Typical forms} & \textbf{Maturity of discovery} \\
\midrule
Factual & discrete facts about entities, events, attributes & KG triples, eventuality graphs, memory entries \cite{bai2025autoschemakg,zhang2020aser,rasmussen2025zep,chhikara2025mem0} & text, KG, tables & high for text sources \\
\addlinespace[4pt]
Conceptual & categories, schemas, relations among facts, models, laws & taxonomies, abstract eventualities, ontologies, equations, rules, alpha factors \cite{wu2012probase,he2024abstract,babaei2023llms4ol,shojaee2025llmsr,luo2023chatrule,tang2025alphaagent} & KG schema, concept graphs, equations, rules, programs & moderate, mostly in niches \\
\addlinespace[4pt]
Procedural & how to perform tasks, methods, algorithms & skill libraries, workflows, APIs \cite{wang2023voyager,wang2024awm,zheng2025skillweaver} & code, workflow graphs, text & moderate, rising fast \\
\addlinespace[4pt]
Metacognitive & knowledge about strategies, applicability, and reliability & insights, reasoning strategies, playbooks \cite{zhao2024expel,ouyang2026reasoningbank,zhang2026ace} & free text, rarely structured & low, conditions rarely recorded \\
\bottomrule
\end{tabularx}
\end{table*}

\subsection{Explicit and implicit knowledge}

We call knowledge explicit when it is encoded in a medium that is separable from the agent that produced it and that can be read, cited, and revised directly, and implicit when it exists only in the parameters or activations of a model. Language models store large amounts of implicit relational knowledge that can be probed by cloze queries \cite{petroni2019lama}, localized in specific computations \cite{meng2022rome,geva2023dissecting}, and sometimes decoded by a linear map \cite{hernandez2024linearity}. Foundation models for tables, graphs, and time series store implicit dependencies that support accurate prediction \cite{hollmann2025accurate,ranjan2026rt,ansari2025chronos2}.

Explicit knowledge affords six properties that implicit knowledge lacks or provides only weakly. It is inspectable, since a person or program can read it. It is editable, since a single unit can be corrected without retraining a model. It is composable, since units can be joined into longer chains of reasoning. It is transmissible, since it can be moved to a different agent, a newer model, or a human. It is verifiable, since it can be tested against new data. And it is attributable, since it can carry provenance. These properties are what accumulation across agents and generations requires, which is why we place explicit knowledge at the center.

Emphasizing explicit knowledge does not mean discarding implicit knowledge. Implicit knowledge remains indispensable for perception, fluent generation, and fast pattern recognition, and the representations and priors of strong models are valuable inputs to knowledge computation (\S\ref{sec:kfm}). What it cannot do is serve as the authoritative, attributable record that others inherit. Our claim is that conversions between the two, and processing within the explicit layer, are where the field is weakest.

Explicit knowledge also has well-known costs. Hand-built knowledge bases such as Cyc required large investments and remained incomplete \cite{lenat1995cyc}. Community-maintained resources such as Wikidata scale better but depend on sustained human curation \cite{vrandecic2014wikidata}. What has changed is that discovery can now be automated, and the main open question is how to keep automatically discovered knowledge useful as it grows.

\subsection{Forms of explicit knowledge}
\label{sec:forms}

Explicit knowledge can be stored in several forms, which differ in what they make easy (Table~\ref{tab:forms}).

\begin{itemize}
\item \textbf{Text and documents} are the most expressive and most widely available form. They support all four types but are hard to query precisely or aggregate over.
\item \textbf{Knowledge graphs} store entities and relations. Schema-defined graphs such as Wikidata use a curated relation vocabulary \cite{vrandecic2014wikidata}. Schema-free graphs induced by LLMs, such as the ATLAS graphs built by AutoSchemaKG from over 50 million documents with more than 900 million nodes and 5.9 billion edges \cite{bai2025autoschemakg}, use relations expressed as open-vocabulary phrases and induce schemas after the fact. Eventuality graphs such as ASER use nodes that are themselves events or states \cite{zhang2020aser}.
\item \textbf{Tables and databases} store facts with fixed attributes and support exact aggregation within a known schema. A table may also hold observations rather than knowledge, in which case it is supporting material.
\item \textbf{Programs, skills, and workflows} store procedural knowledge in executable form \cite{wang2023voyager,zheng2025skillweaver}.
\item \textbf{Equations, rules, and formulas} store conceptual knowledge compactly and precisely \cite{brunton2016sindy,luo2023chatrule}.
\end{itemize}

We do not list parametric encodings, such as adapters generated from documents \cite{charakorn2026d2l} or continuous key-value memories built from triples \cite{wang2025kblam,huang2026atlaskv}, as a form of explicit knowledge. Being derived from explicit material, or regenerable from it, does not make an encoding directly inspectable, citable, or revisable. Such encodings are ways of using knowledge in a model, discussed in \S\ref{sec:use_param}. Likewise, an externally updatable store of continuous vectors differs from a store of readable records with conditions and provenance, even though both can be updated without retraining.

\begin{table}[t]
\centering
\caption{Qualitative comparison of explicit knowledge forms. Aggregation refers to counting, ranking, and comparing over many units. Ratings are our assessment.}
\label{tab:forms}
\vspace{4pt}
\scriptsize
\setlength{\tabcolsep}{2.5pt}
\begin{tabular}{@{}lcccccc@{}}
\toprule
& \rotatebox{70}{Inspectable} & \rotatebox{70}{Precise query} & \rotatebox{70}{Aggregation} & \rotatebox{70}{Noise tolerance} & \rotatebox{70}{Procedural} & \rotatebox{70}{No fixed schema} \\
\midrule
Text & high & low & low & high & medium & high \\
Schema-defined KG & high & high & medium & low & low & low \\
Schema-free KG & high & medium & low & medium & low & high \\
Tables, databases & high & high & high & low & low & low \\
Programs, skills & medium & n/a & n/a & low & high & medium \\
Equations, rules & high & high & n/a & low & low & medium \\
\bottomrule
\end{tabular}
\end{table}

No form dominates, and the same knowledge often has several expressions, for example a rule stated in a paper, encoded as a constraint in a database, and implemented in a program. Aligning and combining such expressions is itself an internalization problem. A knowledge-centric system should therefore keep each carrier in its native structure, with its units, variable roles, relation directions, time stamps, and sources, rather than forcing all knowledge into one form such as a graph or a text serialization.

\section{The Knowledge-Centric Loop}
\label{sec:loop}

Figure~\ref{fig:loop} summarizes the organization we propose. Agent systems $S_1, \dots, S_m$ act in their own environments. They may be run by different organizations, built on different models, and active at different times. An explicit knowledge layer $\mathcal{K}$, a set of units of the form in Equation~\eqref{eq:unit}, connects them through three stages.

\begin{itemize}
\item \textbf{Discovery} $D$ maps interaction traces, documents, observations, and data into new knowledge units, $D: \text{sources} \rightarrow \{\kappa\}$.
\item \textbf{Internalization} consists of two kinds of operators. Transformations $T: \mathcal{K} \rightarrow \mathcal{K}'$ change the layer persistently by conceptualizing, aligning, consolidating, repairing, completing, and versioning units. Reasoning $R: (\mathcal{K}, X, q) \rightarrow (a, e, s)$ takes knowledge, supporting material $X$ such as observations and experience, and a request $q$, and returns a result $a$, which may be a conclusion, a regularity, or an explanatory hypothesis, together with the evidence $e$ it depends on and a status $s$ that says whether it is supported by records, inferred by a model, or a hypothesis awaiting validation. Results worth keeping are written back as new units with that status.
\item \textbf{Use} $U: (\mathcal{K}, \text{task}) \rightarrow \text{agent behavior}$ delivers stored or derived knowledge into a particular agent's decisions, symbolically through context, retrieval, and tools, or parametrically through adapters, key-value memories, and edits.
\end{itemize}

Transformation $T$ and reasoning $R$ interact. A regularity proposed by inductive reasoning enters the layer as a candidate, is validated, and then takes part in further abstraction and reasoning. The boundary between reason $R$ and use $U$ is that $R$ produces knowledge independently of any particular agent, whereas $U$ changes how a particular agent behaves. An agent that calls a reasoning service as a tool is using $U$ to invoke $R$.
\para{Trajectories in the loop.} The three stages can be read through the trajectories that the systems of Figure~\ref{fig:loop} produce continuously. Discovery identifies knowledge candidates in trajectories and other sources. Most of a trajectory remains a record or experience, and a candidate is the part judged worth stating for others, such as a recurring failure together with what resolved it. Internalization forms conditional regularities, strategies, and workflows from specific events and fragments of events, by conceptualizing them (\S\ref{sec:concept}), aligning them with what the layer already holds, and recording their conditions, exceptions, and support. It is not the storage or compression of logs. Use applies the result to later understanding, reasoning, and action, and the feedback and outcomes of that use are new eventualities, which enter discovery again and may confirm, refine, or retract what was used. Trajectories are not the only source, and the loop applies equally to documents and data, but they make the cycle concrete, because the same events can be traced from record to knowledge and back.

We use ``internalization'' for processing that takes place within the knowledge layer, so that knowledge is digested before it is used.\footnote{In Nonaka's SECI model \cite{nonaka1994dynamic}, ``internalization'' denotes conversion of explicit knowledge into tacit knowledge, which in our loop corresponds to parametric use (\S\ref{sec:use_param}). Our internalization stage corresponds more closely to SECI's ``combination'' of explicit knowledge, extended with conceptualization and model-based reasoning over the knowledge layer.}

\para{Across systems.} The layer supports accumulation among systems that do not coordinate on a common task. One system may contribute a finding, a second may test it in a different environment, and a third may use it. Sharing need not be unconditional. Access scopes, permissions, and provenance boundaries can differ by contributor, and the layer can be federated rather than a single central database. What the systems share is a format for units and a discipline for conditions, provenance, and status, not necessarily their data.

\para{Across generations.} Agents, the models that internalize knowledge, and the records themselves have different lifecycles. An agent may be replaced within months, a foundation model may be retrained or replaced, and a record may need to remain valid and interpretable for years. Persistence therefore refers to the continuity of the layer and its records, not to any one model. Inheritance means that a new generation understands, uses, revalidates, and revises inherited knowledge. It also means separating knowledge about the environment from operational advice that depended on an old model or tool, so that a new agent does not inherit its predecessor's limitations.

\subsection{A working objective for internalization}
\label{sec:objective}

Internalization lacks an agreed objective, which is one reason it is rarely studied in its own right (\S\ref{sec:lag}). We propose one as a working formulation. Let $\mathcal{Q}$ be a distribution over the reasoning requests the layer should serve, issued by agents of many systems and of future generations, and including deductive, inductive, and abductive requests that were not anticipated when the layer was processed. Let $u(\cdot)$ score a result of the reasoning operator $R$, accounting for its correctness, its evidence, and whether its status is stated honestly. Let $|\mathcal{K}|$ be a description length of the layer, and let $\mathrm{Inv}(\mathcal{K}, \mathcal{K}')$ hold when a transformation from $\mathcal{K}$ to $\mathcal{K}'$ satisfies the invariants of \S\ref{sec:invariants}. A transformation $T$ from a family $\mathcal{T}$ of operators should then solve
\begin{equation}
\begin{aligned}
\max_{T \in \mathcal{T}} \;\; & \mathbb{E}_{q \sim \mathcal{Q}} \big[\, u\big(R(T(\mathcal{K}), X, q)\big) \big] \;-\; \lambda\, |T(\mathcal{K})| \\
\text{s.t.} \;\; & \mathrm{Inv}\big(\mathcal{K}, T(\mathcal{K})\big),
\end{aligned}
\label{eq:objective}
\end{equation}
where $\lambda \ge 0$ trades usefulness against size. The formulation makes three commitments explicit. Internalization is valuable through the reasoning it enables, so it is judged downstream. Because $\mathcal{Q}$ includes unanticipated requests from parties other than the producer, caching answers is not enough, and abstraction that generalizes is rewarded. And the constraint prevents a smaller layer from being bought by discarding the conditions and evidence that make knowledge reusable. Conceptualization, alignment, consolidation, repair, and forgetting are all instances of $T$ under this objective. We do not claim that Equation~\eqref{eq:objective} is directly optimizable. It states what internalization is for.

\subsection{Maturity and the bottleneck hypothesis}

Table~\ref{tab:maturity} gives our assessment of each stage. Discovery from unstructured text and use by LLMs are well populated with methods and benchmarks. Reliable explicit extraction from structured foundation models, internalization of large and heterogeneous knowledge stores, and use by non-language foundation models are sparsely populated.

We state the central claim as a scoped hypothesis. As the knowledge layer grows in size, in the diversity of its carriers and relation phrasings, in the number of systems contributing to it, and in update frequency, improving internalization yields larger gains per unit of cost than adding more raw knowledge or more context, with discovery and use held fixed. The hypothesis concerns regimes that knowledge-centric agents are entering, namely large, heterogeneous, multi-contributor, and changing stores, and not small curated ones. 

\begin{table*}[t]
\centering
\caption{Maturity of each stage of the knowledge-centric loop, in our assessment of the literature we examined.}
\label{tab:maturity}
\vspace{4pt}
\footnotesize
\renewcommand{\arraystretch}{1.15}
\begin{tabularx}{\textwidth}{@{}P{2.2cm} P{2.4cm} P{2.0cm} Y Y@{}}
\toprule
\textbf{Stage} & \textbf{Sub-area} & \textbf{Maturity} & \textbf{Representative work} & \textbf{Main gap} \\
\midrule
Discovery & Unstructured text & advanced & KG construction \cite{bai2025autoschemakg,zhang2020aser,edge2024graphrag,gutierrez2024hipporag}, IE \cite{wei2023chatie,xu2024genie}, agent memory \cite{park2023generative,rasmussen2025zep,xu2025amem} & noise, conditions and provenance rarely recorded \\
\addlinespace[4pt]
& Experience & advancing & skills \cite{wang2023voyager,zheng2025skillweaver}, workflows \cite{wang2024awm}, insights \cite{zhao2024expel,ouyang2026reasoningbank} & validation, separation from tool-specific advice \\
\addlinespace[4pt]
& Structured data & niche & equations \cite{shojaee2025llmsr,xia2026srscientist}, factors \cite{tang2025alphaagent}, rules \cite{luo2023chatrule}, hypotheses \cite{majumder2025discoverybench,bai2024abductive} & variables usually given, independent validation rare \\
\addlinespace[4pt]
& Explicit extraction from implicit FMs & largely absent & distillation in physics \cite{cranmer2020symbolic,lemos2023orbital}, probing of LMs \cite{meng2022rome} & no general, validated method for tabular, graph, or time series FMs \\
\midrule
Internalization (transformation) & Conceptualization & mature in commonsense, early elsewhere & taxonomies \cite{wu2012probase}, eventuality conceptualization \cite{zhang2022aser,he2024abstract,wang2023cat,wang2024candle} & conditions and evidence across abstraction, other carriers and knowledge types \\
\addlinespace[4pt]
& Consolidation and evolution & early & workflow induction \cite{wang2024awm}, memory evolution \cite{xu2025amem}, playbooks \cite{zhang2026ace} & when to merge, abstract, forget, and version \\
\addlinespace[4pt]
& Repair and completion & mature for curated KGs & link prediction \cite{zhu2021nbfnet,galkin2024ultra}, commonsense population \cite{fang2021ckbp} & open-vocabulary, noisy, temporal, multi-carrier knowledge \\
\midrule
Internalization (reasoning) & Deduction & early & KG-RAG \cite{edge2024graphrag,sun2024tog,luo2024rog}, CLQA \cite{ren2020betae,galkin2024ultraquery,bai2023eventcqa}, text-to-SQL \cite{li2023bird,lei2025spider2} & reliable conclusions under noise and incompleteness, across carriers \\
\addlinespace[4pt]
& Induction & early & symbolic regression and scientific discovery \cite{brunton2016sindy,shojaee2025llmsr,xia2026srscientist,majumder2025discoverybench}, rule mining \cite{galarraga2013amie,luo2023chatrule} & conditions, exceptions, and candidate status kept with induced regularities \\
\addlinespace[4pt]
& Abduction & nascent & abductive NLI \cite{bhagavatula2020abductive}, KG hypothesis generation \cite{bai2024abductive,gao2026controllable,gao2026hypoagent}, hypothesis refinement and falsification \cite{qiu2024phenomenal,wang2024hypothesis,huang2025popper} & multiple explanations with evidence and testable predictions \\
\midrule
Use & Symbolic use & advanced & RAG \cite{lewis2020rag,gao2023ragsurvey}, skills as code \cite{wang2023voyager}, serving systems \cite{yu2024cosmo} & context limits, conflicts \cite{liu2024lost,xie2024chameleon} \\
\addlinespace[4pt]
& Parametric use by LLMs & advancing & editing \cite{meng2023memit,zhang2025conke}, hypernetworks \cite{charakorn2026d2l}, KV memories \cite{wang2025kblam,huang2026atlaskv} & conflicts with priors \cite{cheng2026override}, provenance lost \\
\addlinespace[4pt]
& Non-LLM FMs & largely absent & in-context examples \cite{hollmann2025accurate}, covariates \cite{ansari2025chronos2} & no native interface for conditional rules, constraints, or graphs \\
\bottomrule
\end{tabularx}
\end{table*}

\section{Knowledge Discovery}
\label{sec:discovery}

Knowledge discovery turns what an agent encounters into explicit units. We organize it by three sources: unstructured text, interaction experience, and structured and observational data. The relation discovery literature, which seeks relationships in tables, graphs, time series, and text, is a large part of this stage, and discovery also includes procedural and strategic knowledge distilled from an agent's own experience.

\subsection{From unstructured text: graphs and memories}

\para{Knowledge graph construction.} LLMs perform zero-shot information extraction through multi-turn questioning \cite{wei2023chatie} and in-context relation extraction \cite{wan2023gptre}, a literature reviewed by \cite{xu2024genie}. LLMs also support ontology learning, including term typing and taxonomy discovery \cite{babaei2023llms4ol}. Large-scale extraction is not limited to entities. ASER mines eventualities and their discourse relations from raw text \cite{zhang2020aser}, and FolkScope combines LLM generation with human-in-the-loop annotation to build a knowledge graph of purchase intentions \cite{yu2023folkscope}. AutoSchemaKG removes the need for a predefined schema. It extracts entity and event triples and induces schemas through conceptualization, and its schema induction reaches 92\% semantic alignment with human-crafted schemas without manual intervention \cite{bai2025autoschemakg}. GraphRAG builds an entity graph and community summaries to answer global questions about a corpus \cite{edge2024graphrag}, LightRAG adds graph structure to text indexing with dual-level retrieval \cite{guo2024lightrag}, and HippoRAG combines an LLM-built graph with Personalized PageRank, inspired by the hippocampal indexing theory of memory \cite{gutierrez2024hipporag}. HippoRAG~2 addresses the drop on simple factual tasks seen in earlier structure-augmented RAG \cite{gutierrez2025hipporag2}.

\para{Agent memory.} Agents also extract knowledge from conversational and textual interaction logs. Generative agents store observations and periodically synthesize higher-level reflections \cite{park2023generative}. MemGPT manages memory tiers in analogy to an operating system \cite{packer2023memgpt}, and MemoryBank maintains long-term memory for conversational agents \cite{zhong2024memorybank}. Mem0 extracts and consolidates salient facts from conversations and offers a graph-based variant \cite{chhikara2025mem0}, Zep maintains a temporally aware knowledge graph \cite{rasmussen2025zep}, and A-Mem organizes memories as linked notes following the Zettelkasten method \cite{xu2025amem}. Hierarchical and graph memories organize multi-agent trajectories into insight, query, and interaction levels \cite{zhang2025gmemory} or support active reconstruction of memories during reasoning \cite{ji2026mragent}. These memories are typically private to one system, which is appropriate for personal context but means that what one system learns about the world is not available to others.

\subsection{From experience: procedures and strategies}

Procedural and metacognitive knowledge are discovered from experience. Voyager builds an ever-growing library of executable skills while exploring an open world \cite{wang2023voyager}. Agent Workflow Memory induces commonly reused routines from web navigation trajectories and provides them to the agent for later tasks \cite{wang2024awm}. SkillWeaver lets web agents discover, practice, and distill skills into reusable APIs \cite{zheng2025skillweaver}. ExpeL extracts natural-language insights from collections of experiences \cite{zhao2024expel}, ReasoningBank distills reasoning strategies from both successful and failed experiences \cite{ouyang2026reasoningbank}, and Dynamic Cheatsheet keeps an evolving memory of strategies and code snippets at test time \cite{suzgun2025dc}. Knowledge distilled from experience mixes statements about the environment, such as how a website is organized, with statements about the agent, such as which prompt worked for its model. Only the former should be inherited without revalidation by an agent built on a different model.

\subsection{From structured and observational data: conceptual knowledge}

Relation discovery from structured and observational data yields conceptual knowledge. Symbolic regression recovers governing equations through sparse dictionaries \cite{brunton2016sindy}, physics-inspired decomposition \cite{udrescu2020aifeynman}, or evolutionary search \cite{cranmer2023pysr}, and LLMs now propose equation programs \cite{shojaee2025llmsr}, learn concept libraries \cite{grayeli2024lasr}, or act as agents with analysis tools \cite{xia2026srscientist}. In finance, formulaic alpha factors are discovered by reinforcement learning \cite{yu2023alphagen} and by LLM agents \cite{tang2025alphaagent,li2025rdagentquant}. Logical rules are mined from knowledge graphs \cite{galarraga2013amie} and proposed by LLMs \cite{luo2023chatrule,zhu2023htt}, causal graphs are learned from data \cite{zheng2018notears} and informed by LLM priors \cite{ban2025integrating}, and schema correspondences are found by LLM-based matchers \cite{liu2025magneto}.

Discovery also includes abduction, the inference of plausible explanations for observations \cite{bhagavatula2020abductive}. Over knowledge graphs, abduction can be posed as generating a complex logical hypothesis that explains a set of observed entities, and supervised training alone was found not to guarantee better hypotheses for unseen observations \cite{bai2024abductive,gao2026controllable,gao2026unifying,gao2026hypoagent}. Hypothesis generation from data \cite{zhou2024hypogenic}, data-driven discovery benchmarks \cite{majumder2025discoverybench}, and literature-grounded hypothesis generation in chemistry \cite{yang2025moosechem} are abductive in character, and agentic falsification provides a principled way to validate abduced hypotheses \cite{huang2025popper}. Induction and abduction recur inside internalization (\S\ref{sec:reasoning}). We place them in discovery when they operate on raw data from the environment and in internalization when they operate on the knowledge layer and its supporting records, where their outputs must be reconciled with what is already known.

\subsection{Using implicit capacities versus extracting explicit knowledge}
\label{sec:gap}

A large and growing class of foundation models captures relations implicitly. Tabular in-context learners predict on new tables using priors over structural causal models \cite{hollmann2023tabpfn,hollmann2025accurate,qu2025tabicl}. Relational and graph foundation models transfer across databases and graphs \cite{ranjan2026rt,liu2024ofa,eremeev2025graphpfn}. Knowledge graph foundation models generalize to unseen entities and relations through relation graphs \cite{galkin2024ultra}, prompt graphs \cite{cui2024kgicl}, or prior-data fitted in-context learning \cite{gao2026kgpfn}. Time series models share information across related series and covariates \cite{ansari2025chronos2}, and biological models learn dependencies among genes and cells \cite{cui2024scgpt}.

Two ways of benefiting from these models should be kept apart. Their implicit capacities can be used directly, through their predictions, embeddings, or internal representations, as inputs to further computation (\S\ref{sec:kfm}). This is feasible today, as long as the outputs are treated as model judgments. Their knowledge can also be extracted into explicit units that others can inspect and inherit, and this is where the gap lies. Language models appear to be an exception, since they can be asked to verbalize what they know, but verbalization is not the same as faithful extraction, because a model's stated reasons need not reflect what drives its behavior \cite{turpin2023unfaithful}. For other modalities, existing evidence of extraction is narrow. Symbolic distillation recovers force laws from graph networks trained with strong inductive biases \cite{cranmer2020symbolic} and rediscovered orbital mechanics from solar system data \cite{lemos2023orbital}, and Kolmogorov--Arnold networks can be simplified into formulas \cite{liu2025kan}. Inside language models, relational knowledge can be located and edited \cite{meng2022rome}, and some relations are linearly decodable \cite{hernandez2024linearity}. We found no general method that extracts conditional rules, equations, or graphs from a pretrained tabular or time series foundation model and validates them.

\gapbox{\textcolor{rust}{\textbf{Gap 1.}} Implicit capacities of structured-data foundation models can already be used, but there is, to our knowledge, no validated route from what they capture to reliable explicit knowledge. Extraction methods should query such models under controlled interventions and compile their behavior into rules, equations, or graphs with explicit conditions, and should distinguish fidelity to the model, recovery of the true generating mechanism, and validity under intervention. A rule can be faithful to a model that is itself wrong about the world.}

\subsection{Discovered is not yet validated}

Discovered knowledge enters $\mathcal{K}$ with a status $\rho$ that should reflect its evidence. Three threats are specific to automated discovery. LLMs can emit memorized rather than discovered knowledge, which motivated memorization-resistant benchmarks \cite{shojaee2025llmsrbench}. Explanations attached to discovered artifacts may be unfaithful to how they were produced \cite{jacovi2020faithful,turpin2023unfaithful}. And when many candidates are tried, some will appear valid by chance \cite{simmons2011false,bailey2016backtest}, a problem that grows with adaptive search. Outputs of induction and abduction are candidates by construction and should not be recorded as established knowledge before they are checked. These threats matter more in a knowledge-centric system than in a single study, because unvalidated knowledge that enters the layer will be conceptualized, combined, and reused by other systems and later generations, so errors compound. Recording conditions, provenance, and validation status at discovery time is the cheapest defense.

\section{Knowledge Internalization}
\label{sec:internal}

Internalization is everything that happens to knowledge between discovery and use. We argue that it is the least developed and most consequential stage of the loop.

\subsection{Internalization as a persistent object}
\label{sec:why}\label{sec:lag}

A knowledge layer that only grows becomes less useful over time. Automatically discovered knowledge is redundant across words and carriers, noisy from accumulated extraction errors, incomplete, inconsistent as sources disagree or facts change, and overly specific to single instances; moreover, most of its value lies in combinations of units rather than single records. Human knowledge institutions spend much of their effort on this processing: writing reviews and textbooks, merging procedures into standards, retracting errors, and answering questions that no single source answers. Today AI systems mostly do this ad hoc: an LLM agent retrieves material, aligns and combines it in context for its current task, and discards the intermediate structure when the session ends. The next agent, or the same agent after a model upgrade, starts again from raw records.

\pos{\textcolor{navy}{\textbf{Core claim.}} Knowledge internalization should not only be work that each agent performs for its current task. It should be a persistent object of training and maintenance, independent of any single agent. Knowledge foundation models should learn representation, semantic alignment, condition-preserving abstraction, compositional reasoning, and revision across carriers, so that these capacities and their results serve different systems and generations instead of being rebuilt from all historical material at every task and every model upgrade.}

This claim does not assume that LLMs cannot solve complex tasks; strong LLMs, structured-data foundation models, and symbolic executors can all take part in internalization. Rather, it distinguishes completing one knowledge-intensive task from maintaining reusable understanding, an agent's transient context from an inheritable layer, and on-the-fly alignment from a shared capacity for alignment and composition. In terms of complementary learning systems theory, where the hippocampus stores new episodes rapidly while the neocortex slowly consolidates them into structured knowledge \cite{mcclelland1995cls}, most current agent systems implement only the fast store (or borrow hippocampal indexing for retrieval \cite{gutierrez2024hipporag}) and omit slow consolidation.

Why has internalization lagged? First, LLM agents can bypass it on small stores and simple questions by placing retrieved records directly in context. Second, its value is realized downstream and often by other parties, lacking a task-independent objective (which Equation~\eqref{eq:objective} addresses). Third, benchmarks measure extraction quality or end-task accuracy rather than whether operators preserve conditions and evidence status. Fourth, knowledge stores remain fragmented across projects rather than forming a shared pretraining substrate. Fifth, extraction pipelines rarely record the provenance and versioning needed for evolution. We next examine the three components of internalization: transformations $T$, reasoning $R$, and knowledge foundation models.

\subsection{Transformations and operator invariants}
\label{sec:concept}

Transformations $T: \mathcal{K} \rightarrow \mathcal{K}'$ reorganize and maintain the knowledge layer through cross-carrier alignment, condition-preserving conceptualization, repair and completion, and operator invariants.

\para{Aligning knowledge across carriers.} Knowledge relevant to a question often spans papers, measurement tables, knowledge graphs, rules, and programs. Converting everything into a single graph or text serialization drops native structure—units and variable roles in tables, relation directions in graphs, quantifiers in rules, and preconditions in programs. Alignment instead establishes which elements across carriers refer to the same entities, variables, concepts, or claims, recording correspondences with confidence and provenance (e.g., schema matching \cite{liu2025magneto}, entity linking, and grounding text to eventuality graphs \cite{jiayang2024eventground}), while attaching supporting observations and model judgments without promoting them to facts.

\para{Condition-preserving conceptualization.} Conceptualization abstracts specific instances into higher-level concepts and regularities that apply to new situations \cite{wang2025conceptsurvey}.\footnote{We use conceptualization for mapping instances to concepts, complementing ontological conceptualization \cite{gruber1993translation}, whose commitments (such as keeping roles distinct from types) check proposed abstractions via invariant I3.} It is the core transformation of internalization because it turns what happened into what tends to happen across entities, event arguments, procedure parameters, table variables, and trajectory fragments. Given instance units $\kappa_1, \dots, \kappa_n$ sharing a pattern and a concept map $\phi$ (e.g., an isA taxonomy), conceptualization forms $\hat\kappa = (\hat c, f, \hat C, \hat\pi, \hat\rho)$ where (i) $\hat c$ replaces instances by concepts under $\phi$ while preserving argument roles and step dependencies, (ii) $\hat C$ records shared applicability conditions and exceptions from counter-instances, (iii) $\hat\pi$ links to each supporting $\kappa_i$ so instance detail remains recoverable, and (iv) $\hat\rho$ counts independent sources rather than mentions and marks $\hat\kappa$ as a candidate until validated. The inverse operation, instantiation, applies $\hat\kappa$ to a new instance $x$ when $\phi(x)$ matches and $\hat C$ holds.

\para{Scale and neighboring routes.} In commonsense acquisition, conceptualization has scaled from taxonomic mapping in Probase \cite{wu2012probase} and eventuality graphs in ASER \cite{zhang2022aser} to neural verification over ATOMIC \cite{he2024abstract}, joint contextualized conceptualization and instantiation in CAT and CANDLE \cite{wang2023cat,wang2024candle}, and schema induction in AutoSchemaKG \cite{bai2025autoschemakg}; benchmarks such as AbsPyramid show that LLMs struggle with such abstraction zero-shot but improve with targeted training \cite{wang2024abspyramid}. Beyond commonsense, neighboring lines of work perform closely related operations: event schema induction mines typical event chains and hierarchies \cite{chambers2008narrative,li2023schema}; agent workflow and memory methods distill trajectories into routines, APIs, strategies, and evolving playbooks \cite{wang2024awm,zheng2025skillweaver,ouyang2026reasoningbank,zhang2026ace,xu2025amem}; and program library learning trades library size against solution compression \cite{ellis2020dreamcoder,grand2024lilo}, akin to Equation~\eqref{eq:objective}. However, iterative context rewriting suffers from brevity bias and context collapse that erode domain detail \cite{zhang2026ace}, and existing schemas or agent memories rarely maintain explicit conditions, exceptions, independent-source counts, and derivation links across systems.

\para{Repair, completion, and evolution.} While link prediction and completion are mature on curated KGs \cite{yao2019kgbert,zhu2021nbfnet,galkin2024ultra,xie2026ngdbzoo}, and commonsense population \cite{fang2021ckbp}, temporal graphs \cite{rasmussen2025zep}, and human curation \cite{vrandecic2014wikidata} address parts of maintenance, multi-contributor discovered knowledge requires entity resolution across carriers, open-vocabulary completion that keeps inferred items distinct from observed ones, provenance-weighted conflict resolution, and versioned deprecation that preserves revision history for later generations.

\para{Invariants for internalization operators.}\label{sec:invariants} Recording $C$, $\pi$, and $\rho$ is insufficient if automated operators corrupt them (e.g., merging ten copies of one report into ten apparent supports). Beyond provenance vocabularies, operators themselves should satisfy five checkable invariants, distinct from downstream predictive accuracy:
\begin{enumerate}
\item[\textbf{I1}] \textbf{Provenance is never lost.} Every derived unit, alignment, and stored result links to the units and supporting records it was derived from.
\item[\textbf{I2}] \textbf{Support counts independent sources.} Duplicates and units that share a source do not increase support or confidence.
\item[\textbf{I3}] \textbf{Roles, types, and units are compatible.} A merge, alignment, or abstraction never unifies elements that play different roles, have incompatible types, or are measured in incompatible units or at incompatible granularity.
\item[\textbf{I4}] \textbf{Conditions, exceptions, and status are not silently dropped.} An abstraction may widen its conditions only when evidence covers the wider scope, known exceptions are carried forward, and inferred or hypothesized content is never relabeled as observed.
\item[\textbf{I5}] \textbf{Retraction propagates.} When a source is retracted or a unit is refuted, every unit and stored result derived from it is flagged or revised.
\end{enumerate}

\subsection{Reasoning over noisy knowledge}
\label{sec:reasoning}\label{sec:noisy}

\begin{figure*}[t]
\centering
\includegraphics[width=0.92\textwidth]{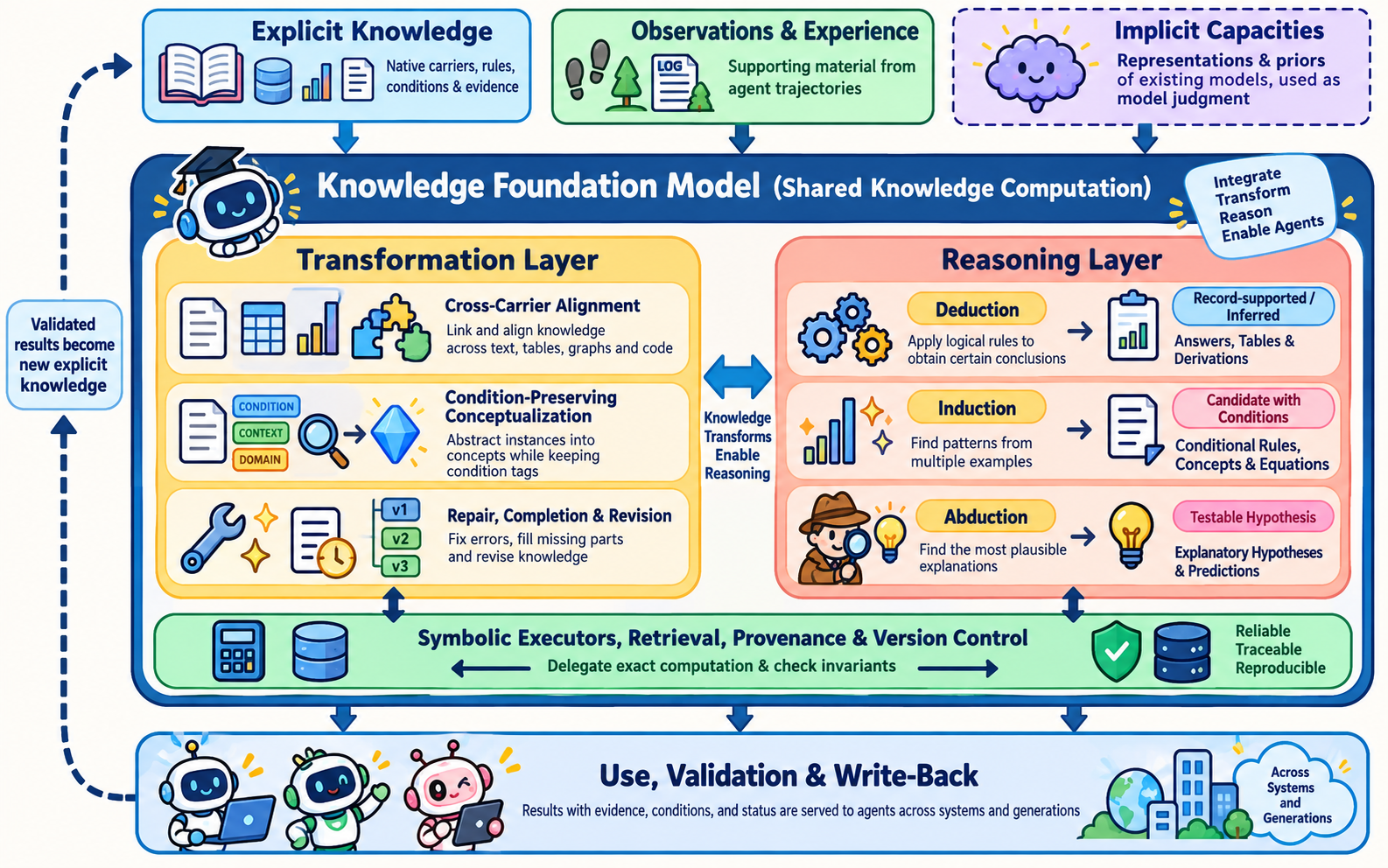}
\caption{A knowledge foundation model inside the knowledge layer. It combines explicit knowledge, supporting observations and experience, and the implicit capacities of existing models, organizes shared computation across transformation and reasoning layers while delegating exact computation to symbolic components, and produces deductive, inductive, and abductive results whose status is recorded before they are written back. Implicit priors inform inference but do not become knowledge without evidence.}
\label{fig:kfm}
\end{figure*}

Much of the value of a knowledge layer lies in results that no single stored unit contains. We organize the reasoning operator $R$ by three classical forms of inference (Table~\ref{tab:reasoning}), as a long-term view of the capabilities internalization needs rather than a claim that a single network head provides all three.

\begin{table}[t]
\centering
\caption{Three forms of reasoning over the knowledge layer.}
\label{tab:reasoning}
\vspace{4pt}
\scriptsize
\setlength{\tabcolsep}{2.5pt}
\begin{tabularx}{\columnwidth}{@{}P{1.25cm} Y Y Y@{}}
\toprule
\textbf{Form} & \textbf{Question} & \textbf{Input $\rightarrow$ output} & \textbf{Status of output} \\
\midrule
Deduction & What follows from given knowledge, rules, and conditions? & knowledge from several carriers and a query $\rightarrow$ answers, result tables, derivations, evidence & supported by records if premises are, otherwise inferred \\
Induction & What generalizable regularity do many instances support? & supporting instances and observations $\rightarrow$ conditional rules, concepts, candidate equations & candidate, with conditions and exceptions \\
Abduction & What hypothesis would explain an observation? & anomaly or phenomenon and existing knowledge $\rightarrow$ candidate explanations, evidence, testable predictions & hypothesis, possibly several \\
\bottomrule
\end{tabularx}
\end{table}

\para{Deduction.} Deduction derives what follows from given knowledge, rules, and conditions, including multi-hop traversal, cross-carrier joins, set operations, and aggregation. In an automatically accumulated layer, neither direct in-context reasoning nor translating requests into symbolic queries suffices on its own: LLMs do not reliably execute complex joins and aggregations over large structured stores in context, while text-to-SQL or Cypher pipelines remain brittle on dirty, enterprise-scale schemas (e.g., 40.1\% accuracy for ChatGPT on BIRD \cite{li2023bird} and 21.3\% for an o1-preview agent on Spider~2.0 \cite{lei2025spider2}). More fundamentally, even when query translation succeeds, symbolic execution is exact only with respect to the visible records: missing facts silently drop objects from joins and counts, unresolved duplicates inflate aggregates, and open-vocabulary phrasings fail to match query predicates. Deductive reasoning must therefore combine exact execution where records are complete with model-based completion where they are not, explicitly separating record-supported steps from inferred estimates.

\para{Induction.} Induction proposes generalizable regularities—rules, concepts, governing equations, and scientific laws—from accumulated instances and observations, as in symbolic regression, scientific discovery, and rule mining \cite{brunton2016sindy,shojaee2025llmsr,xia2026srscientist,galarraga2013amie,luo2023chatrule}. Because finite instances and potentially biased coverage do not logically guarantee a generalization, induced regularities must carry applicability conditions, known exceptions, and candidate status (P2), which is where induction and conceptualization meet.

\para{Abduction.} Abduction generates and refines explanatory hypotheses for anomalies or observations \cite{bhagavatula2020abductive,bai2024abductive,gao2026hypoagent,qiu2024phenomenal,wang2024hypothesis,gao2026controllable,gao2026unifying}. Because several explanations may fit the available records, abductive reasoning must return supporting evidence and testable predictions to distinguish competing candidates until they are validated or falsified \cite{huang2025popper}.

\para{Existing routes and their limits.} Across all three forms, $R$ must keep record-supported conclusions, model inferences over missing facts, and hypotheses awaiting validation apart, reporting intervals or abstaining when missing data render an answer non-identifiable. Existing composite and neural routes each address part of this space (Table~\ref{tab:routes}): probabilistic databases model uncertain tuples \cite{dalvi2007probdb}; neural databases execute select--project--join operators over short text facts \cite{thorne2021neuraldb}; and neural graph databases and complex logical query answering (CLQA) pair latent representations with neural query engines to handle missing edges and open-vocabulary semantic units \cite{hamilton2018gqe,ren2020query2box,ren2020betae,ren2023ngdb,galkin2024ultraquery,bai2025agenticngdb,li2026ngdbench}. Yet no single route combines cross-carrier alignment, explicit conditions, calibrated aggregation, and unified deductive, inductive, and abductive inference—directly motivating a shared knowledge computation model.

\begin{table*}[t]
\centering
\caption{Routes to reasoning over knowledge and what each leaves open in the setting of \S\ref{sec:noisy}, in our assessment. Each route is a candidate component of a knowledge foundation model rather than a competitor to it.}
\label{tab:routes}
\vspace{4pt}
\footnotesize
\renewcommand{\arraystretch}{1.2}
\begin{tabularx}{\textwidth}{@{}P{3.4cm} Y Y@{}}
\toprule
\textbf{Route} & \textbf{What it does well} & \textbf{What remains open for heterogeneous, noisy, incomplete knowledge} \\
\midrule
LLM over retrieved context \cite{lewis2020rag,gao2023ragsurvey} & open semantics, flexible reasoning, evidence in context & stable aggregation over large sets, uneven use of long context \cite{liu2024lost} \\
Graph retrieval and global summarization \cite{guo2024lightrag,gutierrez2024hipporag,sun2024tog,luo2024rog,jiang2023structgpt,edge2024graphrag} & local paths and neighborhoods, and corpus-level sensemaking through community summaries in GraphRAG & exact, auditable set statistics, estimates of what is missing \\
Neural databases over text \cite{thorne2021neuraldb} & schema-free SPJ queries over short text facts with symbolic aggregation & cross-carrier joins, explicit conditions and provenance, separating inferred from observed facts \\
Neural graph databases and CLQA \cite{hamilton2018gqe,ren2020query2box,ren2020betae,arakelyan2021cqd,wang2023lmpnn,wang2021efo1,yin2024fit,bai2023eventcqa,ren2023ngdb,galkin2024ultraquery,bai2025agenticngdb} & multi-hop logical queries under missing edges, zero-shot transfer to new graphs in UltraQuery & counting and aggregation, robustness to spurious edges, duplicates, paraphrases, carriers other than graphs \\
Knowledge graph foundation models \cite{galkin2024ultra,cui2024kgicl,gao2026kgpfn} & relations represented without per-relation parameters & evaluation mostly on curated vocabularies and link prediction \\
LLM-generated queries with symbolic execution \cite{li2023bird,lei2025spider2} & exact computation over the given records & grounding of open phrasings, invisibility of missing and duplicated facts, error propagation across modules \\
Cleaning, resolution, probabilistic databases \cite{dalvi2007probdb} & principled handling of known error types and uncertain tuples & open-vocabulary alignment across carriers, induced and abduced results \\
Structured-data foundation models \cite{hollmann2025accurate,ranjan2026rt,ansari2025chronos2} & strong priors for prediction on tables, databases, series & conditioning on explicit knowledge, explicit outputs (Gaps 1 and 3) \\
\bottomrule
\end{tabularx}
\end{table*}

\subsection{Knowledge foundation models}
\label{sec:kfm}

These considerations motivate knowledge foundation models (KFMs): models, or model-centered systems, that provide shared neural capacity for knowledge computation inside the knowledge layer (Figure~\ref{fig:kfm}).

\para{A knowledge foundation, not a data-task foundation.} A KFM is defined by knowledge computation rather than by a single data modality or collection of task heads. It learns representation, cross-carrier alignment, condition-preserving abstraction, and deductive, inductive, and abductive reasoning that transfer zero-shot to unseen knowledge stores contributed by independent systems, much as UltraQuery transfers across unseen graphs \cite{galkin2024ultraquery}. Serving agents across systems and generations as well as non-agent analytical systems, a KFM has a lifecycle separate from both the agents and the explicit records (P1), and delegates complete, exact computation to symbolic executors while propagating uncertainty over candidate mappings, missing links, and evidence into downstream inference.

\para{Building on implicit capacities without overriding records.} A KFM can draw on the embeddings, representations, and pattern priors of pretrained LLMs and structured-data foundation models \cite{liu2024ofa,kim2024carte,ranjan2026rt}, treating their predictions as model judgments rather than observed facts. Implicit priors can propose alignments or missing relations, but model confidence is not evidence quality, compressed vectors cannot replace explicit roles, conditions, and provenance, and priors must not silently override conflicting explicit records. Whether a shared capacity for cross-carrier alignment, condition-preserving abstraction, and reasoning can be acquired by pretraining—building on directions in large knowledge models \cite{chen2024lkm,huang2026lkm}, relational and KG foundation models \cite{ranjan2026rt,galkin2024ultraquery,gao2026kgpfn}, and neural graph databases \cite{ren2023ngdb,bai2025agenticngdb}—is the central open question we advocate.

\gapbox{\textcolor{rust}{\textbf{Gap 2.}} To our knowledge, no existing approach, whether a single neural model or a composite of retrieval, parsing, and execution, provides shared, transferable capacity for aligning knowledge across native carriers, abstracting it with conditions, and reasoning deductively, inductively, and abductively over it while keeping record-supported conclusions, model inferences, and hypotheses apart under noise and incompleteness. Building such knowledge foundation models, together with transformation operators that keep the layer useful as it grows across systems and generations, is the central research priority we advocate.}

\section{Knowledge Use}
\label{sec:use}

Knowledge serves agents through two channels, which mirror the explicit and implicit distinction of \S\ref{sec:knowledge}.  In a knowledge-centric setting the consumer is often not the producer. It may run a different model, have different tools and permissions, or belong to a later generation, so the interface must convey conditions, provenance, and status along with content.

\subsection{Symbolic use}

The simplest channel places knowledge in the model's context. Retrieval-augmented generation retrieves passages and conditions generation on them \cite{guu2020realm,lewis2020rag,borgeaud2022retro}, a large literature surveyed in \cite{gao2023ragsurvey}. Graph-based retrieval extends this to structured knowledge \cite{edge2024graphrag,gutierrez2025hipporag2}. Using an eventuality graph for a text task first requires grounding the text to the graph, which EventGround treats as a problem in its own right for narrative reasoning \cite{jiayang2024eventground}. At production scale, COSMO generates e-commerce commonsense knowledge and serves it to applications at Amazon \cite{yu2024cosmo}. Procedural knowledge is used by execution: skills are called as code \cite{wang2023voyager}, APIs are invoked \cite{zheng2025skillweaver}, and workflows guide action generation \cite{wang2024awm}. Metacognitive knowledge is used as instructions, such as retrieved strategies \cite{ouyang2026reasoningbank} or playbooks \cite{zhang2026ace}.

Symbolic use keeps knowledge separable and attributable in the store, but three limitations remain, and a fourth concerns what attribution actually achieves. Context is costly and finite. Models use long contexts unevenly, with performance often depending on where relevant information appears \cite{liu2024lost}. Context can conflict with what the model believes. LLMs can be highly receptive to coherent external evidence that contradicts their parametric memory, yet they show confirmation bias when the evidence is partly consistent with it \cite{xie2024chameleon}, so the effect of supplied knowledge is hard to predict. Finally, the fact that cited evidence was placed in context does not show that the answer depends on it, and a retraction in the store does not reach answers already produced unless invariant I5 is enforced end to end.

\subsection{Parametric use}
\label{sec:use_param}

A second channel converts explicit knowledge into parameters or parameter-like memories \cite{huang2026inparameter}. Knowledge editing writes individual facts into model weights \cite{meng2022rome} and scales to thousands of edits \cite{meng2023memit}. ConceptEdit adds conceptualization and instantiation to the editing pipeline, so that edits about commonsense knowledge extend beyond the literal statement \cite{zhang2025conke}, and CoCoLM teaches pretrained language models complex commonsense by training them over the eventuality graph ASER \cite{yu2022cocolm}. Parametric RAG converts documents into parameters for test-time knowledge enhancement \cite{su2025prag,tan2025dyprag}. Hypernetworks generate adapters in a single forward pass, from a task description in Text-to-LoRA \cite{charakorn2025t2l}, from a context in Generative Adapter \cite{chen2024genadapter}, and from a document in Doc-to-LoRA, which stores needle-in-a-haystack information well beyond the target model's native context length \cite{charakorn2026d2l}. Key-value approaches encode knowledge base entries as continuous key-value pairs that the model attends to directly. KBLaM integrates more than ten thousand triples into an 8B model with an 8K context window \cite{wang2025kblam}, and AtlasKV scales this idea to knowledge graphs with about one billion triples using less than 20GB of GPU memory \cite{huang2026atlaskv}. Memory layers add trainable key-value lookups that provide large capacity for factual knowledge at low compute cost \cite{berges2024memorylayers}, and nearest-neighbor language models interpolate with an external datastore \cite{khandelwal2020knnlm}.

Parametric use removes context costs, but it gives up some of what makes knowledge explicit. Four limitations stand out. First, injected knowledge competes with pretrained priors. For Doc-to-LoRA, accuracy on facts that contradict pretraining collapses to 46.4\% on the deepest conflicts, and it falls from 68\% to 16\% as the base model's prior on the contradicted fact grows stronger \cite{cheng2026override}. Second, edits do not propagate to logically related facts \cite{cohen2024ripple}. Third, parametric knowledge can be directional, since models trained on ``A is B'' do not automatically answer ``B is A'' \cite{berglund2024reversal}, and knowledge is reliably extractable only if it was seen in diverse forms during training \cite{allenzhu2024physics31}. Fourth, once knowledge is in parameters, its provenance and conditions are no longer attached to it, so it cannot be cited, audited, or cleanly retracted. Parametric encodings are also tied to one model, so a new generation must regenerate them from the explicit records, which is another reason the records, not the encodings, must remain authoritative (P1). Key-value methods keep knowledge separable and updatable \cite{wang2025kblam,huang2026atlaskv}, which makes them a promising middle ground for use, although their continuous entries are not readable records.

\subsection{Beyond language models}

Both channels exist almost exclusively for language models. Foundation models for structured data accept some side information, but not explicit knowledge in the sense of \S\ref{sec:knowledge}. A tabular in-context learner accepts labeled rows as context, and its prior over dependencies is fixed at pretraining \cite{hollmann2023tabpfn,hollmann2025accurate}. Time series models accept related series and covariates \cite{ansari2025chronos2}, which is the closest existing mechanism, but not statements such as ``demand drops on public holidays'' or ``these two sensors are on the same road''. Graph and relational models use text embeddings of node features or column names \cite{liu2024ofa,kim2024carte,ranjan2026rt}, which conveys semantics but not rules, constraints, or known equations. Practitioners can approximate such knowledge by engineering features or by constraining outputs after prediction, but these workarounds are task-specific and cannot be retracted or audited as knowledge. As a result, knowledge discovered by agents, and the large stores of scientific and domain knowledge accumulated by humans, have no general route into the many structured-data tasks that these models serve. A KFM that reasons over the layer and passes conditioned results to such models is one possible bridge, but it does not remove the need for native interfaces.

\gapbox{\textcolor{rust}{\textbf{Gap 3.}} Non-language foundation models lack a native interface through which explicit knowledge, such as conditional rules, constraints, causal graphs, or equations, can be supplied at inference time, can stably affect predictions, and can later be withdrawn. Such an interface should let correct knowledge help, let incorrect or inapplicable knowledge degrade predictions gracefully, and ensure that withdrawn knowledge stops having an effect.}

\subsection{Where a persistent knowledge layer matters}
\label{sec:settings}

Two settings show what use looks like when knowledge comes from operations and from people's lives rather than from documents, and where its value lies.

\para{Enterprise operations.} Enterprise data platforms already maintain a shared operational representation. In Palantir Foundry, for example, the ontology is described as a digital twin of the organization, in which object types define real-world entities or events, link types define relationships between them, and action types define sets of edits that a user can make at once. Successful action submissions are recorded in an action log as objects linked to the objects they changed, which the documentation presents as supporting audits, analysis of patterns across decisions, and decision workflows. Read through our loop, business data and operation records are sources and experience, the curated business representation shared by an organization's applications belongs to its knowledge layer, and decisions and their execution are use whose outcomes become new records. These correspondences describe roles in the knowledge lifecycle. They are not a mapping between the capabilities of such a platform and our three stages. The research question they raise is how to form, from operation records and their outcomes, knowledge that carries conditions and provenance and can be revised, such as the hypothetical regularity when a sole-source supplier is delayed by more than a few days, planners switch to an alternative carrier, except for high-value orders, which go to manual approval, together with the independent cases that support it and the outcomes that followed. In this setting an automatically induced regularity should be a candidate for review by the people accountable for the process, not an executable rule that takes effect without validation, and human overrides of suggested decisions are a valuable source of exceptions. We do not assess here which of these capabilities existing platforms already provide.

\para{Personal agents and consumer services.} Personal agents, companionship and social assistants, and e-commerce services draw on three kinds of knowledge that should be kept apart. General commonsense, such as the typical stages of planning a wedding, is largely available in the parameters of current LLMs, and a separate resource for it adds little. Individual event state, such as where a particular user is in such a process, what they have said they do not want, and which plans mentioned in past conversations remain open, is in no model and accumulates over months. Population statistics over private behavioral data, such as how purchases shift in the months after an engagement in a given region, exist only in a platform's logs and support calibrated estimates where an LLM can only guess. The last two are naturally sequences of eventualities, and conceptualization turns them into conditional regularities, such as a user who does not want work reminders in the evening, rather than the over-general claim that the user dislikes talking about work. They also have to be managed continuously. Personal knowledge links to products, services, and other people, so links need permissions, and an abstraction should not be visible to someone who could not see the instances that support it. Users must be able to inspect, correct, and delete what an agent holds about them, and corrections and deletions must reach derived units (I5) and be honored when an agent forgets. Production systems such as COSMO already generate e-commerce commonsense knowledge from user behavior and serve it to applications \cite{yu2024cosmo}.

\para{What the value consists of.} In both settings, the value of an explicit layer is not only, and not necessarily, higher accuracy on a single answer, since a strong model given the same raw records may answer as well. It lies in reuse of the same knowledge across tasks, products, and parties, in lower cost when many requests are served from processed knowledge instead of reprocessing raw records, in auditability of what a decision or recommendation relied on, and in continuity when the model behind an agent or service is replaced. Enterprise and consumer settings thus support the same claim: experience produced by different systems needs to become knowledge that can be reused, governed, and revised as it evolves.

\section{A Research Agenda}
\label{sec:agenda}

We translate the position into seven challenges. They are not independent. The common knowledge unit (C2) defines the object on which transformation operators (C3, C5) act, knowledge foundation models (C4) reason over their output, and knowledge layers across systems and generations (C6) are where all of them meet. Extraction from and interfaces for non-language foundation models (C1, C7) connect the loop to the models that serve structured-data tasks.

\para{C1. Explicit extraction from implicit foundation models.} Develop methods that extract rules, equations, or graphs from tabular, graph, and time series foundation models, with explicit conditions, and that distinguish fidelity to the model, mechanism recovery, and validity under intervention (Gap~1). Synthetic data with known generating processes, which several of these models already use for pretraining \cite{hollmann2023tabpfn,ansari2025chronos2,eremeev2025graphpfn}, provides ground truth.

\para{C2. A common knowledge unit across carriers.} Adopt representations that record content, form, conditions, provenance, and status (Equation~\eqref{eq:unit}) for every carrier in its native structure, together with alignments between carriers, so that knowledge produced by different agents can be merged, validated, and versioned.

\para{C3. Conceptualization and consolidation operators with invariants.} Define conceptualize, align, merge, forget, and version operators for each knowledge type and carrier, building on entity and eventuality conceptualization \cite{zhang2022aser,he2024abstract,wang2024candle} and extending it to fragments of agent trajectories and operation records, tables, programs, procedural knowledge, and metacognitive knowledge. Characterize them by the invariants of \S\ref{sec:invariants}, not only by downstream accuracy, and make explicit how each decides what to keep, how far to abstract, and when to merge, refine, or retract (\S\ref{sec:objective}).

\para{C4. Knowledge foundation models.} Build models, possibly neural-symbolic, that learn shared capacity for alignment, condition-preserving abstraction, and deductive, inductive, and abductive reasoning over heterogeneous, noisy, and incomplete knowledge, that build on the representations and priors of existing language and structured-data models without letting priors override records, that delegate exact computation, and that report the status of each result (Gap~2). Openly available stores such as the ATLAS graphs \cite{bai2025autoschemakg} and eventuality graphs \cite{zhang2020aser}, together with structured databases, can serve as training and study material.

\para{C5. Knowledge repair and evolution at scale.} Extend entity resolution, completion, population, and conflict resolution to automatically discovered knowledge from many contributors and carriers \cite{fang2021ckbp}, keep completed items distinct from observed ones, and support temporal evolution with retained history \cite{rasmussen2025zep}, so that the layer improves rather than degrades as it grows.

\para{C6. Knowledge layers across systems and generations.} Design layers to which independent agent systems contribute and from which they draw, with scoped sharing, permissions, and provenance boundaries rather than a requirement of one central store, analogous in spirit to the role Wikidata plays for human editors \cite{vrandecic2014wikidata}. Ensure that abstractions do not reveal instances their readers are not permitted to see, and that corrections and deletions requested by the people a record concerns propagate (\S\ref{sec:settings}). Support inheritance by later generations of agents and of KFMs, including revalidation of inherited knowledge, separation of knowledge about the environment from model-specific operational advice, and interpretation of records by models other than the one that wrote them.

\para{C7. Knowledge interfaces for all foundation models.} Give non-language foundation models interfaces for conditioning on explicit knowledge (Gap~3), and make parametric use by LLMs robust to conflicts with priors \cite{cheng2026override} while keeping knowledge separable and retractable.

\para{A key comparison.} Across these challenges, one comparison tests whether persistent knowledge processing is worth its cost. The baseline is an agent built on the same model with access to the same raw records, such as trajectories, logs, or operation histories, together with retrieval and code execution, under the same budget of tokens and tool calls. The question is what a processed knowledge layer adds to it. We expect any increment to appear in reuse across instances and tasks, in the handling of conditions and exceptions, in updates and retractions that reach dependent conclusions, and in transfer to agents built on another model family, rather than necessarily in accuracy on a single question. The comparison does not presume that explicit knowledge is more accurate, and finding that the increment appears only in some of these dimensions, or in some regimes and not others, would sharpen the position.

\section{Alternative Views}
\label{sec:alt}

\para{``Scale is enough.''} One view holds that larger models trained on more data will internalize whatever knowledge is needed, making explicit knowledge unnecessary. We agree that scale improves implicit knowledge, and a KFM should exploit it. But implicit knowledge cannot be moved to a new model, cited, or selectively corrected, and studies of parametric knowledge show directional storage \cite{berglund2024reversal}, dependence on training-data diversity \cite{allenzhu2024physics31}, and incomplete propagation of edits \cite{cohen2024ripple}. More fundamentally, knowledge that many independent agents discover after a model is trained must go somewhere, and retraining a model for each discovery is neither economical nor auditable. Each new model generation would also start without the record of why earlier conclusions were revised.

\para{``Long context makes knowledge stores unnecessary.''} If context windows grow large enough, one could place all relevant knowledge in context. Context remains costly, models use it unevenly \cite{liu2024lost}, and aggregation over millions of facts is beyond any practical window. Context also ends with the session and belongs to one system, so it does not accumulate across systems or generations.

\para{``Each agent can internalize on demand.''} A capable LLM agent can retrieve, align, and reason over material for each task, so a separate internalization stage may seem redundant. For a single task this can work well. The cost is that the work is repeated by every agent and after every model upgrade, its intermediate results are not kept for others, and its conclusions are not maintained when sources change. Our claim is about where this work should live, not about whether LLMs can do it. An LLM-based system that performs it persistently and for other parties is one realization of what we advocate. The strongest form of this view is an agent with raw records, retrieval, and code execution, and the comparison at the end of \S\ref{sec:agenda} is meant to test it.

\para{``Symbolic knowledge has failed before.''} Expert systems and hand-built knowledge bases struggled with the cost of acquisition and with brittleness \cite{lenat1995cyc}. Our position differs in two ways. First, acquisition is now largely automatic and has been shown at scale for taxonomies \cite{wu2012probase}, eventualities \cite{zhang2020aser}, and schema-free graphs with billions of edges \cite{bai2025autoschemakg}, and automatically generated knowledge is served in production systems \cite{yu2024cosmo}. Second, we do not propose symbolic reasoning alone. We propose learned models and operators that act on explicit knowledge and build on implicit capacities, combining the robustness of learned representations with the properties of explicit storage.

\para{``A database with a good parser is enough.''} For curated, complete stores with stable schemas, it may be, and symbolic execution should then do the exact computation. Our concern is the distance between exact execution and reliable knowledge conclusions (\S\ref{sec:noisy}). When records are missing, duplicated, phrased in open vocabularies, conflicting, or biased in coverage, an exact answer over the records can be a wrong answer about the world. Cleaning, resolution, and probabilistic databases address parts of this, and they are components of the systems we advocate. What they do not yet provide is shared capacity for alignment across carriers and for inductive and abductive results with explicit status.

\para{``Internalization is just indexing.''} One might argue that internalization reduces to better retrieval engineering. Retrieval selects existing units, whereas internalization changes the knowledge layer by aligning, conceptualizing, merging, completing, and correcting, and it produces conclusions, regularities, and explanations that are not stored anywhere. 

\section{Conclusion}
\label{sec:conclusion}
Knowledge accumulates because people write down what is worth keeping, and others, in their own time and in later generations, refine, combine, and build on it. AI agents are now built by many independent parties and replaced within a few months, yet what they learn stays mainly in parameters, in context, and in private stores that are appended to but rarely processed. We have argued that the next generation of AI should be knowledge-centric. Agents of many systems discover explicit knowledge, a knowledge layer independent of any single agent internalizes it through conceptualization, alignment, maintenance, and deductive, inductive, and abductive reasoning, and agents of other systems and later generations use the result through symbolic and parametric channels. Agent trajectories, operation records, and personal event histories make this concrete: their eventualities are experience, condition-preserving conceptualization of events and event fragments turns them into reusable knowledge, and new outcomes revise it. LLM agents have made rapid progress on discovery from text and on use by language models. The gaps lie in reliable extraction from and use by structured foundation models, and above all in internalization. We hope the community will treat internalization as a persistent object of training and maintenance, and knowledge foundation models, whose foundation is knowledge and its computation and which build on the capacities of existing models, as first-class research problems.

\bibliographystyle{abbrv}
\bibliography{references}

\end{document}